\documentclass[lettersize,journal]{IEEEtran}

\usepackage{cite}
\usepackage{amsmath,amssymb,amsfonts}
\usepackage{algorithmic}
\usepackage{graphicx}
\usepackage{textcomp}
\usepackage{xcolor}
\def\BibTeX{{\rm B\kern-.05em{\sc i\kern-.025em b}\kern-.08em
    T\kern-.1667em\lower.7ex\hbox{E}\kern-.125emX}}

\usepackage{hyperref}
\usepackage[linesnumbered,ruled]{algorithm2e}
\usepackage[super]{nth}
\usepackage{float}
\usepackage{wrapfig}
\usepackage{caption}
\usepackage{subcaption}
\usepackage{svg}
\usepackage{booktabs}

\usepackage{amsmath}
\usepackage{amssymb}
\usepackage{mathtools} 
\usepackage{bm}

\newcommand{\lie}[1]{\mathcal{L}_{#1}}

\newcommand{\state}[0]{\bm{x}}
\newcommand{\nstate}[0]{n}
\newcommand{\action}[0]{\bm{u}}
\newcommand{\naction}[0]{m}

\newcommand{\ostate}[0]{\bm{x}_o}
\newcommand{\nostate}[0]{d}
\newcommand{\dostate}[0]{\dot{\bm{x}}_o}

\newcommand{\motion}[1]{\bm{\kappa}_{#1}}
\newcommand{\motiont}[2]{\bm{\kappa}_{#1}({#2})}

\begin{document}

\title{Proactive Local-Minima-Free Robot Navigation: Blending Motion Prediction with Safe Control}

\author{Yifan Xue$^{1\dagger}$, Ze Zhang$^{2,\dagger}$, Knut Åkesson$^{2}$ and~Nadia Figueroa$^{1}$~\IEEEmembership{Member,~IEEE}
\thanks{Manuscript received September 16, 2025; Revised December 22, 2025; Accepted February 6, 2026.}
\thanks{This work is supported in part by the National Science Foundation (NSF) Foundational Research in Robotics (FRR) program under NSF CAREER Award Grant (No. FRR-2443721), and in part by the AIHURO project (Vinnova 2022-03012). The computations were enabled by resources provided by the National Academic Infrastructure for Supercomputing in Sweden (NAISS), partially funded by the Swedish Research Council (grant agreement no. 2022-06725). }
\thanks{$^\dagger$These authors contributed equally to this work.} 
\thanks{$^{1}$Y. Xue and N. Figueroa are with the Department of Mechanical Engineering and Applied Mechanics, University of Pennsylvania, Philadelphia, PA 19104 USA. {\tt\footnotesize \{yifanxue, nadiafig\}@seas.upenn.edu}.}
\thanks{$^{2}$Z. Zhang and K. Åkesson are with the Department of Electrical Engineering, Chalmers University of Technology, SE-412 96 Gothenburg, Sweden. {\tt\footnotesize \{zhze, knut.akesson\}@chalmers.se}.}
}

\markboth{IEEE ROBOTICS AND AUTOMATION LETTERS. PREPRINT VERSION. ACCEPTED February, 2026}%
{Xue, Yifan and Zhang, Ze \MakeLowercase{\textit{et al.}}: Proactive Local-Minima-Free Robot Navigation: Blending Motion Prediction with Safe Control}


\maketitle

\begin{abstract}
This work addresses the challenge of safe and efficient mobile robot navigation in complex dynamic environments with concave moving obstacles. Reactive safe controllers like Control Barrier Functions (CBFs) design obstacle avoidance strategies based only on the current states of the obstacles, risking future collisions. To alleviate this problem, we use Gaussian processes to learn barrier functions online from multimodal motion predictions of obstacles generated by neural networks trained with energy-based learning. The learned barrier functions are then fed into quadratic programs using modulated CBFs (MCBFs), a local-minimum-free version of CBFs, to achieve safe and efficient navigation. The proposed framework makes two key contributions. First, it develops a prediction-to-barrier function online learning pipeline. Second, it introduces an autonomous parameter tuning algorithm that adapts MCBFs to deforming, prediction-based barrier functions. 
The framework is evaluated in both simulations and real-world experiments, consistently outperforming baselines and demonstrating superior safety and efficiency in crowded dynamic environments.
\end{abstract}

\begin{IEEEkeywords}
Dynamic obstacle avoidance, nonconvex obstacles, learning-based control barrier function
\end{IEEEkeywords}

\section{Introduction}
\IEEEPARstart{T}he ability to handle uncertain and dynamic environments is crucial for robots to exhibit intelligence and adaptability. With the growing demand for Autonomous Mobile Robots (AMRs) in both industrial transportation and daily service tasks, navigation environments have become increasingly complex and dynamic. Robots are often required to operate alongside humans in shared spaces without physical barriers, posing significant challenges to ensuring safety and efficiency. While motion planning and obstacle avoidance in static environments are well-established, solving dynamic obstacle avoidance is still an unsolved problem~\cite{mavrogiannis_2023_socialnavi}. 
In particular, autonomous systems face three challenges in dynamic multi-obstacle avoidance: (i) sudden motion changes of the obstacles, (ii) future feasibility of the trajectories, and (iii) concavity of the environment. Moving obstacles can form concave unsafe regions when intersecting or overlapping with one another, which often leads to the robot freezing or stopping near the obstacle boundaries. For dynamic obstacles, their states and motions may change rapidly, undergoing continuous rotation and translation. Recent efforts to account for sudden motion changes in dynamic obstacle avoidance include increasing algorithm computation speed to make the controller reactive, and planning trajectories using receding-horizon control methods using motion predictions of the dynamic obstacles. Yet, the applicability of such techniques does not scale well to multi-obstacle scenarios like humans dynamically forming concave obstacle regions.

\begin{figure}[t]
    \centering
    \includegraphics[width=0.99\linewidth]{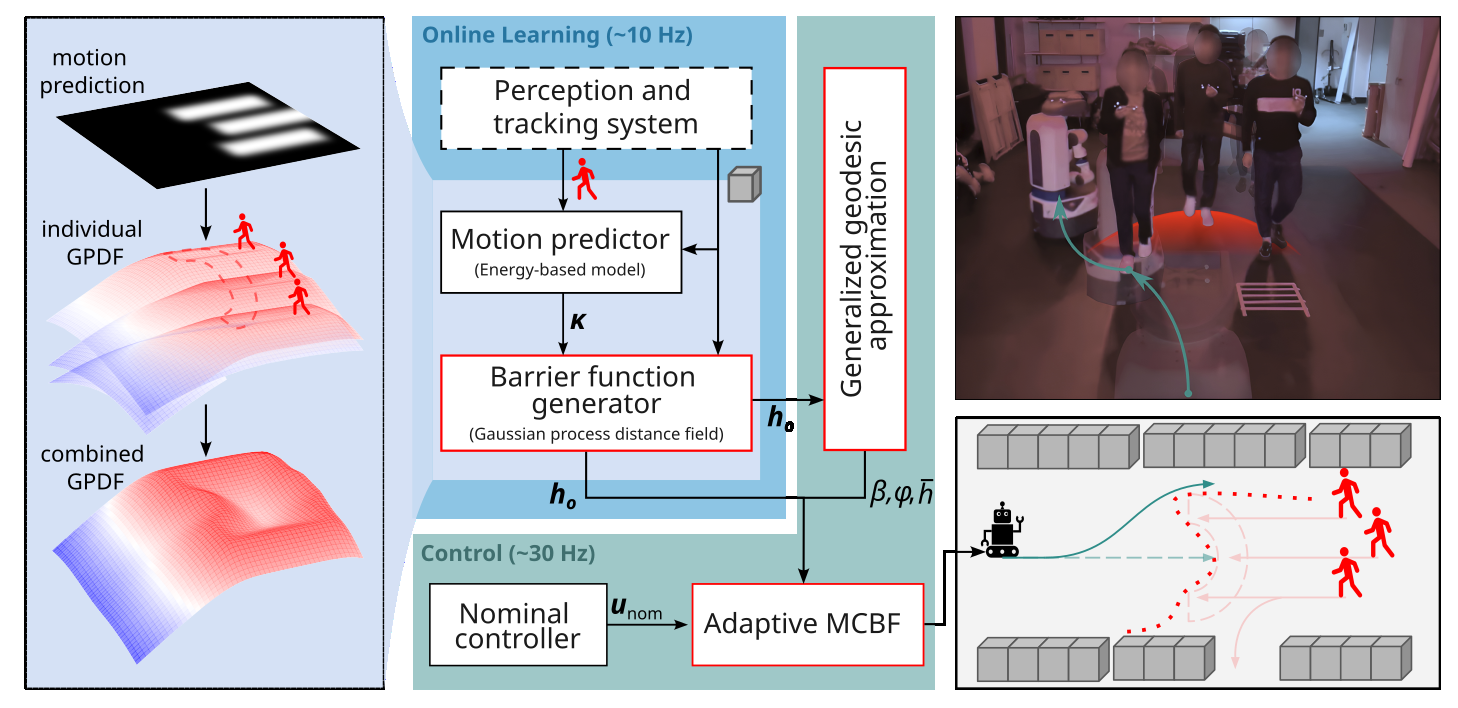}
    \caption{General pipeline of the proposed MMP-MCBF approach. Given a perception and tracking system, a motion predictor makes predictions of dynamic obstacles, which are converted into barrier functions via online learning using Gaussian Process Distance Fields (GPDFs). The proposed adaptive MCBF controller utilizes the barrier functions to generate safe and efficient control actions. In this work, the learning-based motion predictor is pre-trained.}
    \label{fig:concept}
\end{figure}

Reactive control is a type of control scheme that directly responds to the environment or sensory inputs without either a complicated modeling and understanding of the environment or extensive planning~\cite{correll_2022_introduction, billard_2022_learning}. Existing state-of-the-art reactive safe controllers, including Control Barrier Function-based Quadratic Programs (CBF-QPs)~\cite{ames_2019_cbf, cosner2023robust, adaptivecbf} and dynamical system modulation~\cite{huber_2022_avoiding, billard_2022_learning}, focused on minimally altering the nominal control actions and only circumnavigate the obstacles when they are close enough to threaten robot safety. Recent advances, such as on-Manifold Modulation \cite{onManifoldMod} and Modulated CBF-QPs (MCBF) \cite{xue_2025_mcbf}, show the ability to plan trajectories among concave obstacle regions without freezing the robot. Despite their real-time performance, such non-proactive strategies have no guarantees for the future feasibility of the robot. 
Multiple dynamic obstacles can easily corner a robot if it fails to react during the initial time steps, when escape from the surroundings is still feasible.
To enhance robot future safety using CBF-QPs, in~\cite{prednomcbf}, the future distance to obstacles is leveraged when tracking its nominal trajectory. However, it requires accurate knowledge of obstacle states in the future time steps, which is nearly impossible in practice. In~\cite{predstatecbf}, predictive CBF is formulated using dynamic obstacles' future reachable spaces, approximated as ellipses. Such an approximation is over-conservative and can lead to unnecessary deadlocks when feasible paths to circumnavigate the obstacles can still be found. 
Due to these challenges, many works have focused on adopting Model Predictive Control (MPC) and Model Predictive Path Integral (MPPI) for dynamic obstacle avoidance \cite{zhang_2023_wtampc, Heuer_2023_proactive, samavi_2024_sicnav, mppi}. Although such receding-horizon methods can naturally integrate motion prediction to ensure robot feasibility and safety in the next couple of time steps, they face challenges in terms of computational efficiency and handling non-convex obstacles \cite{mpccbf, mpccbflayered}. Alternatively, Reinforcement Learning (RL) methods are also explored for mobile robot navigation \cite{tai_2017_sim2real, ceder_2024_ddpgmpc}. They can learn complex policies from raw sensor data and store the decision-making process, which handles complex environments and guarantees real-time performance. However, they often suffer from issues such as training instability, lack of interpretability, and difficulty in generalizing to unseen environments. Moreover, ensuring safety during training and deployment remains a critical challenge, especially in dynamic environments~\cite{Xu_2023_drllimit}.

In this work, we propose a novel framework that integrates Multimodal Motion Prediction (MMP) with the safe controller MCBF-QP to enable safe and efficient mobile robot navigation in dynamic environments with non-convex obstacles. Both model-based Constant Velocity Models (CVMs)~\cite{scholler_2020_cvm} and learning-based Energy-Based Models (EBMs)~\cite{ze_2025_ebm} are used as motion predictors. The framework’s two main technical contributions, detailed in Section \ref{sec:method}, are: the online learning of barrier functions from motion predictions, and the adaptive on-manifold safe control algorithm. We validate the proposed method in both simulation and real-world experiments, demonstrating superior performance in dynamic environments.

\section{Problem Formulation}
\label{sec:problem}

Given control-affine nonlinear dynamics, $f:\mathbb{R}^{\nstate} \to \mathbb{R}^{\nstate}$ and $g:\mathbb{R}^{\nstate}\to \mathbb{R}^{\nstate \times \naction}$, with the state $\state \!\in\! \mathbb{R}^{\nstate}$ and the control input $\action \!\in\! \mathbb{R}^{\naction}$, for a continuous-time system $\dot{\state} = f(\state) + g(\state)\action$, the following concepts for safe control can be defined.

\textbf{Safe Control:} In an environment with a set $O$ of occupied unsafe regions, for an unsafe region $\ostate \in \mathbb{R}^{\nostate}$ where $o\in O$, a continuously differentiable function $h_o\!:\mathbb{R}^\nstate \!\times\! \mathbb{R}^{\nostate} \rightarrow \mathbb{R}$ is a barrier function if the safe set $C_o$ (outside the obstacle), boundary set $\partial C_o$ (on the obstacle's boundary), and unsafe set $\neg C_o$ (inside the obstacles) are defined~\cite{ames_2019_cbf}:
\begin{align}
    C_o &= \{\state \in \mathbb{R}^\nstate, \ostate \in \mathbb{R}^{\nostate}: h_o(\state, \ostate)>0\}, \label{eq:safe_region_o} \\
    \partial C_o &= \{\state \in \mathbb{R}^\nstate, \ostate \in \mathbb{R}^{\nostate}: h_o(\state, \ostate)=0\}, \label{eq:boundary_o} \\
    \neg C_o &= \{\state \in \mathbb{R}^\nstate, \ostate \in \mathbb{R}^{\nostate}: h_o(\state, \ostate)<0\}. \label{eq:unsafe_region_o}
\end{align}
The barrier function $h_o(\state, \ostate)$ can be a safety measurement acquired by elevating the Distance Function (DF) to the robot state space, where the DF $s_o(\bm{\xi}, \ostate)$ is defined as the orthogonal distance from a robot position $\bm{\xi}$ to the boundary of the unsafe set $o$~\cite{oleynikova_2016_signed}. 
While $h_o(\state, \ostate)$ commonly incorporates the DFs between the robot and the \emph{physical} obstacle boundaries \cite{xue_2025_mcbf}, the barrier functions in this work also measure the distance from the robot to the \emph{virtual} region of predicted human reachable spaces. This allows the proposed framework to strategically circumvent corridors that are soon to be blocked by human activities, leading to safer and more efficient trajectory generation. 

\textbf{Estimated Forward Reachable Set:} Given the initial state $\ostate(0) \in \mathbb{R}^{\nostate}$, estimated possible motion set $U_o$, and dynamics $f_o(\ostate,\bm{u}_o)$ of a moving obstacle, its Estimated Forward Reachable Set (EFRS) at time $\tau>0$ is all possible regions where the dynamic obstacle can reach given $\tau$ seconds, i.e.,
\begin{align}
    \hat{\mathcal{R}}(\tau) = \{\ostate(\tau) | &\exists \bm{u}_o(t) \!\in\! U_o, \nonumber \\
    &\dot{\bm{x}}_o(t) = f_o(\ostate(t), \bm{u}_o(t)) \,, t \!\in\! [0,\tau]\}.
\end{align}
Given the inherent uncertainty in human motion, an EFRS should include possible multimodality and local randomness.

\textbf{Assumptions:} In this work, assumptions are made for two aspects: perception and prediction. For perception, the static layout of the environment is known and mostly stays unchanged. Point cloud representations of all obstacle surfaces within a distance $b$, without considering sensor visibility limitations, are accessible. In open areas, such information can be obtained from onboard multi-sensor systems, such as LiDARs and cameras. In occluded scenarios, an omniscient perception system \cite{zhang_2023_wtampc} can be employed.

For motion prediction, to ensure general accuracy, all dynamic obstacles move rationally and purposefully. In this work, we employ reactive motion predictors that do not explicitly model interactions between robots and humans. 
Although interaction-coupled predictors can capture richer dynamics in dense environments, they can be seamlessly integrated into our framework; we leave this extension to future work.
We also assume that sufficient data are available for learning-based motion predictors to capture most motion patterns.

\section{Preliminaries}
\label{sec:preliminaries}
This work addresses two sub-problems: modeling groups of obstacles using Gaussian processes informed by motion prediction and controlling AMRs for reactive obstacle avoidance.

\subsection{Predictive Unsafe Region}
Let the current time be 0. Given a dynamic obstacle $o$ and a motion predictor $G_{\text{MP}}$, its estimated future motions are,
\begin{equation}
    \motiont{}{\tau} = G_{\text{MP}}(\motiont{e}{0},\tau),
\end{equation}
where $\motiont{e}{0}$ is the extensive motion information available at the current instant, $\tau\in(0,\tau_{\max}]$ is the predictive time offset, and $\motiont{}{\tau}$ is its future motion at time $\tau$. The actual $\motiont{e}{0}$ depends on the chosen predictor. The predicted motion is then converted to a unified potentially occupied area, which can be regarded as the EFRS of the obstacle. The motion predictor $G_{\text{MP}}$ is not necessarily the same as the obstacle's dynamics $f_o$, but they can serve the same purpose of generating the EFRS.

A common way to make predictions is based on the historical motion profile of the obstacle. For instance, the CVM assumes that a moving object keeps its motion tendency with a constant velocity. An EFRS can be generated by inflating the predicted trajectory of the CVM, as shown in Fig. \ref{fig:wrap sim rep}. While CVMs are simple but effective in many scenarios \cite{scholler_2020_cvm}, to adapt to more complex environments, a data-driven EBM~\cite{lecun_2006, ze_2025_ebm} is also used in this work, which can generate multimodal motion prediction over multiple time steps in one shot,
\begin{equation}
    \langle \motiont{}{\tau}|\tau=1,2,\ldots,\tau_{\max} \rangle=G_{\text{EBM}}\left(\motiont{e}{0}\right).
    \label{eq:mp_ebm}
\end{equation}
For the EBM motion predictor, $\motiont{e}{0}$ is the current and multiple previous positions of the obstacle plus the bird's-eye-view environmental map, and $\motiont{}{\tau}$ is a probability map of the obstacle's future position at step $\tau$. In the case of CVMs, $\motiont{e}{0}$ includes only the current and past positions of the object.

\subsection{Gaussian Process Distance Field} 
Given a set of boundary points $\mathcal{P}=\{\bm{p}_i\}^M_1$ of obstacles, a Gaussian Process Distance Field (GPDF)~\cite{gpdf, le_2023_gpdf} can be established to estimate a continuous and differentiable distance function to obstacles regarding any point in the workspace.  
Define a latent field $o(\bm{p}) \sim \mathcal{G}(0, k_o(\bm{p}, \mathcal{P}))$ following a Gaussian process with the covariance kernel $k_o$, and the inverse function $f_{\text{inv}}$ mapping to the distance field $f_{\text{dist}}(\bm{p})$,
\begin{align}
    f_{\text{inv}}\left(k_o(\bm{p}, \mathcal{P})\right) &\coloneq ||\bm{p}-\mathcal{P}||, \\
    f_{\text{dist}}(\bm{p}) &= f_{\text{inv}}(o(\bm{p})).
\end{align}
Let $K_o^{\mathcal{P}}$ be the covariance kernel of given obstacle points, and $k_o^{\bm{p}}\!=\!k_o(\bm{p}, \bm{p})$, according to Gaussian process regression,
\begin{align}
    \bar{o}(\bm{p}) &= k_o(\bm{p}, \mathcal{P}) \underbrace{\left(K_o^{\mathcal{P}}+\sigma_o^2 \mathbf{I}\right)^{-1}\cdot\bm{1}}_{\alpha(\mathcal{P}, \sigma_o)}, \\
    \mathbf{cov}(o(\bm{p})) &= k_o^{\bm{p}} - k_o(\bm{p}, \mathcal{P})\left(K_o^{\mathcal{P}}+\sigma_o^2 \mathbf{I}\right)^{-1}k_o(\mathcal{P},\bm{p}),
\end{align}
where $\sigma_o$ is the noise covariance, $\bm{1}$ is a vector of ones (let $k_o(\mathcal{P}, \mathcal{P})\!=\!\bm{1}$), and $\alpha(\cdot)$ is the learned model. The gradient is,  
\begin{equation}
    f_{\text{grad}}(\bm{p}) = \nabla_{\bm{p}} f_{\text{dist}}(\bm{p}) = \frac{\partial f_{\text{inv}}}{\partial o}  \left[
    \nabla_{\bm{p}}k_o(\bm{p},\mathcal{P})  \alpha(\mathcal{P}, \sigma_o)
    \right].
\end{equation}
Based on boundary points of obstacles, GPDFs provide both distance fields and gradient fields in a unified form. 

\begin{figure}[t]
    \centering
    \includegraphics[width=0.99\linewidth]{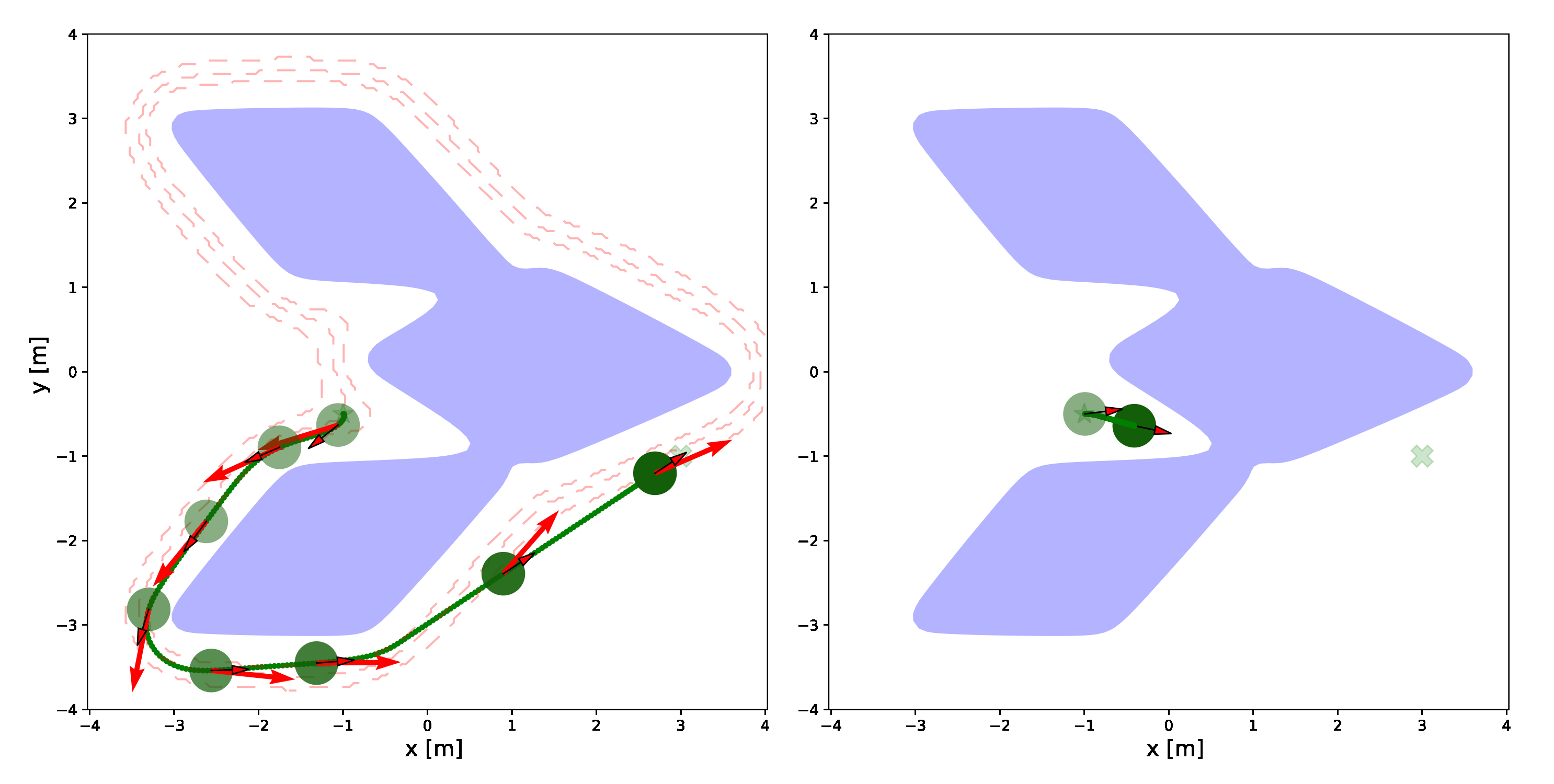}
    \vspace{-5mm}
    \caption{Modulated CBF (left) vs. CBF (right). When facing a complex concave obstacle (in purple), modulated CBFs generate an additional exit ``force'' (long red arrows) introduced by Eq.~\eqref{eq:mod_phi_cbf_constraint_affine}, guiding the robot along the isoline (dashed lines) direction $\phi(\state, h_o)$ and away from the obstacle, whereas standard CBFs become trapped in local minima.}
    \label{fig:comp_cbf_mcbf}
\end{figure}

\subsection{Modulated Control Barrier Functions}
Many safe reactive controllers, including the standard CBF-QPs, suffer from local minima or saddle points~\cite{onManifoldMod,LukesDS,clfcbfEquilibria}. 
To generate safe inputs that ensure no local minima, we use the modulation based on-Manifold Modulated CBF-QP (MCBF-QP)~\cite{xue_2025_mcbf}. Control inputs synthesized by CBF-QPs inevitably result in undesired stable equilibria even for fully-actuated robots~\cite{clfcbfEquilibria,xue_2025_mcbf}, which occur when $\dot{\state}=0$. A robot trapped at an undesired equilibrium will reduce the efficiency of a planned trajectory and even prevent it from reaching the goal.

\begin{figure*}[t]
    \centering
    \includegraphics[width=\linewidth]{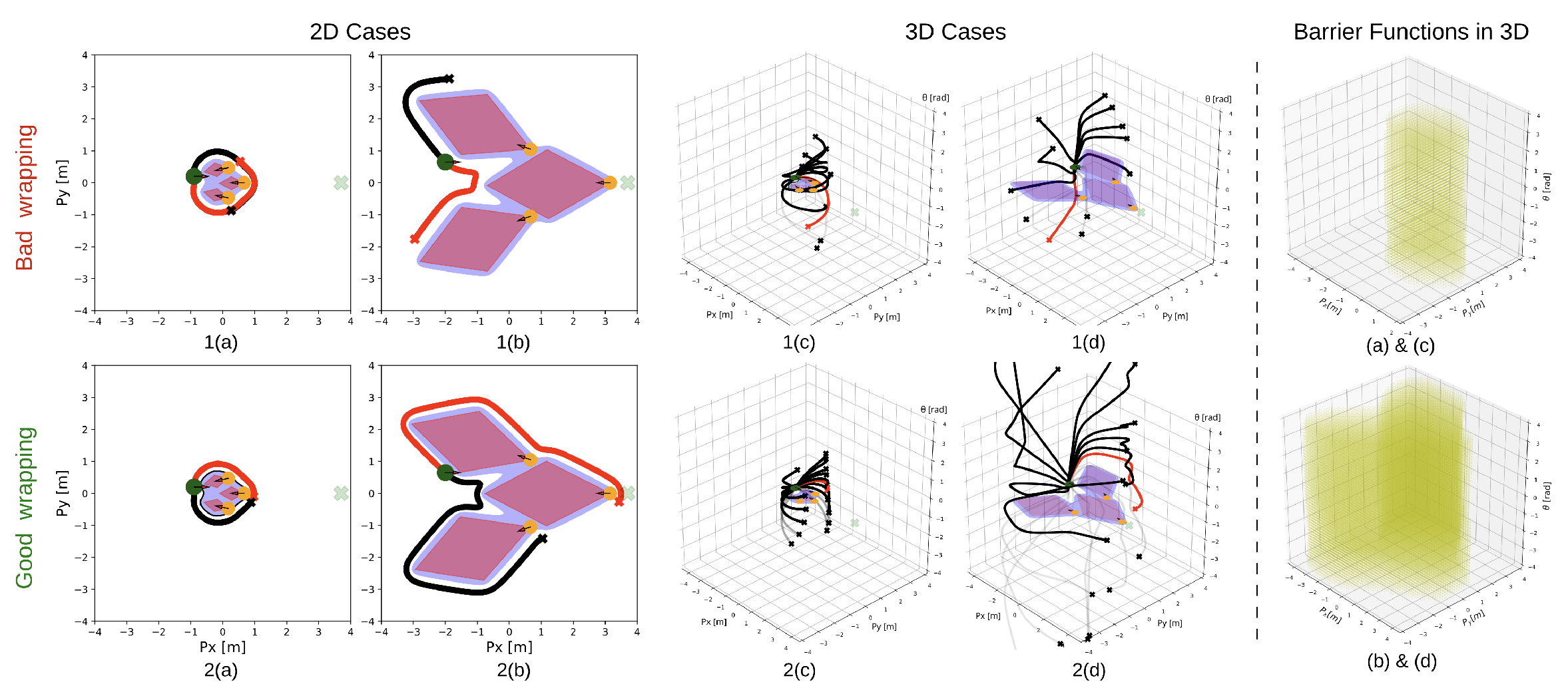}
    \vspace{-5mm}
    \caption{Performance of tangent vector selection using geodesic approximation with constant $\beta$ (first row until the dashed divider, $\beta=0.02$ in 2D, $\beta=0.05$ in 3D) versus generalized geodesic approximation with adaptive $\beta$ selection (second row until the dashed divider). Pink polygons are the estimated forward reachable space of the three humans from the motion predictor under fixed 0.05s time steps and 60 iterations. The last column is the visualization of the augmented CBFs used in the 3D cases. \label{fig:wrap sim rep}}
\end{figure*}

To eliminate undesired equilibria of CBF-QP–type solvers, MCBF-QP modulates the dynamics and projects components onto the tangent planes of barrier functions, introducing tangential velocities. This is achieved by introducing an additional constraint on the standard CBF-QP formulations as in~\eqref{eq:mod_phi_cbf_constraint_affine}. 
\begin{align}
    \action_{\text{mcbf}} &= \arg\min_{\action \in \mathbb{R}^\naction}\;(\action-\action_\text{nom})^\top(\action-\action_\text{nom}), & \nonumber\\
    \text{s.t.}\quad
    &\lie{f} h_o + \lie{g} h_o \action + \nabla_{\ostate}h_o^\top\dostate \geq - \alpha \left(h_o\right),\label{eq:cbf-qp safety constraints}\\
    &\phi(\state,h_o)^\top f(\state)+\phi(\state,h_o)^\top g(\state)\action \geq \gamma, \label{eq:mod_phi_cbf_constraint_affine}
\end{align}
where $h_o=h_o(\state, \ostate)$ is the control barrier function in the context of CBF. $\lie{f}$ and $\lie{g}$ are the Lie derivatives of $h_o$, so $\lie{f}h_o = \nabla_{\state}h_of(\state)$ and $\lie{g}h_o= \nabla_{\state}h_og(\state)$, where $\nabla_{\state}h \in \mathbb{R}^{\nstate \times 1}$ can be viewed as the normal direction from the robot state $\state$ to the unsafe set surface. Here, $\gamma>0$ is a tuning parameter, and the tangent direction $\phi(\state, h_o)$ can be derived using various obstacle exit strategies, such as hessian or geometric approximation~\cite{xue_2025_mcbf, onManifoldMod}. 

MCBF-QP ensures the same safety guarantee as the standard CBF-QP through the use of the CBF constraint in  Eq.~\eqref{eq:cbf-qp safety constraints}. Given a consistent normalized tangent velocity guidance vector $\phi(\state, h_o)$ parallel to $h_o$'s isolines, and a proper $\gamma$, Eq.~\eqref{eq:mod_phi_cbf_constraint_affine} ensures circumvention of the obstacle $h_o$, i.e., local-minimum-free. This is because $\phi^\top\dot{\state}\geq \gamma$ will keep moving the robot in one direction along the isolines until reaching the side that is closer to the target, as demonstrated in Fig. \ref{fig:comp_cbf_mcbf}.

\section{MMP-MCBF Framework}
\label{sec:method}
To leverage the strengths of motion predictors and the ability of safe reactive controllers to navigate concave environments, we propose a proactive and adaptive MMP-MCBF framework, which constructs virtual DFs based on the EFRSs of moving obstacles and continuously updates these functions through online learning, as depicted in Fig. \ref{fig:concept}.

\subsection{Barrier Function Online Learning with Motion Prediction}
\label{sec:barrier function construction}
Different motion prediction methods output predictions $\motion{}$ in different formats. To make them compatible with CBF-QPs, we transform motion prediction into unified EFRSs $\hat{\mathcal{R}}$ of obstacles. In our online learning pipeline, the obtained EFRSs at each iteration are fed into GPDFs to learn DFs, which are then converted into control barrier functions.
To obtain a GPDF, boundary points $\mathcal{P}$ with proper density are required. 

For CVM-based motion predictors, the basic idea is to define an information process space \cite{rios_2015_proxemics} as the EFRS and sample boundary points from it. Based on the estimated velocity of a moving obstacle, a predicted trajectory can be generated. By symmetrically inflating this trajectory about its centerline, a rhombus-shaped region is constructed, which serves as the information-processing space, such as in Fig. \ref{fig:wrap sim rep}. In this work, CVM-based motion predictors are used in the simulated social navigation task and real-world experiments. 
For EBMs, more complicated postprocessing is needed since they generate complex multimodal motion prediction in the form of discrete probability maps. The pixel values of these maps represent the probability of occupancy. According to Eq.~\eqref{eq:mp_ebm}, a stack of probability maps for all predictive time steps is produced for each query. A uniform occupancy map $\motion{\text{occ}}$ over all time steps is obtained by summing all probability maps, i.e., $\motion{\text{occ}}\!=\!\sum_{\tau=1}^{\tau_{\max}} \motiont{}{\tau}$. By setting a threshold $\delta_{\kappa}$, the probabilistic occupancy map is converted into a binary occupancy map. A boundary polygon without self-intersection can be obtained through edge detection and contour detection, i.e., $\text{CONTOUR}(\text{EDGE}(\motion{\text{occ}}, \delta_{\kappa}))$, which encloses the estimated reachable set. 
Specifically, the function $\text{EDGE}(\motion{\text{occ}}, \delta_{\kappa})$ converts $\motion{\text{occ}}$ into a binary edge map according to $\delta_{\kappa}$.
By interpolating this contour and feeding the result into an online learning GPDF, we obtain the DF $s_o(\cdot)$ for the learned EFRS, from which the control barrier function $h_o(\state, \ostate)$ can be computed.

\subsection{Adaptive Modulated Control }
\label{sec:adaptive MCBF}
Standard MCBF-QP \cite{xue_2025_mcbf} is designed to circumvent obstacles, given the constraint in Eq.~\eqref{eq:mod_phi_cbf_constraint_affine}. In general, the shortest path circumventing a single obstacle may not be the best one in dense multi-moving-obstacle environments. Also, MCBF-QP is designed for non-deforming obstacle avoidance applications and cannot adapt to constantly changing EFRSs. To resolve the greedy path-planning limitation and ensure stable performance in the presence of deforming unsafe sets, we propose two improvements that render MCBF-QP proactive and adaptive.

\paragraph{Generalized Geodesic Approximation} During the navigation, the virtual obstacles defined as the EFRS for pedestrians can overlap with other physical obstacles in the environment, thereby blocking certain passageways to the target. When such scenarios occur, computing the exit strategy $\phi(\state, h_o)$ with respect to either the predicted reachable space or nearby physical obstacles may lead the robot to freeze at the intersecting region. Therefore, whenever a predicted reachable space of a dynamic obstacle is close enough to the robot, we instead generate $\phi(\state, \Bar{h})$ using the combined barrier function from all obstacles within distance $b$ to the robot. Given a set of obstacles $Q = \{o | h_o(\state, \ostate)\leq b, o\in O\}$, the combined single obstacle representation can be computed~\cite{onManifoldMod} as 
\begin{equation}
    \Bar{h}(\state) = -\frac{1}{\rho}\log\left(\sum_{o \in Q} \exp{(-\rho [h_o(\state, \ostate)+1])}\right)-1,
    \label{eq:uni h}
\end{equation}
where $\rho$ is a positive user-selected constant. The smaller the value of $\rho$, the smoother the edges in the distance field. 
The resulting function $\Bar{h}$ can represent multiple discrete obstacles using a single function. 

Given the combined barrier function $\Bar{h}$, the obstacle's exit direction $\phi(\state,\Bar{h})$ can be obtained. Let $\bm{e}^{(0)}$ be one out of $m$ uniformly sampled candidate directions in $\mathbb{R}^{\nstate}$, satisfying the requirement in \eqref{eq:e requirement} and subject to $\bm{e}^{(0)} \notin \mathcal{N}(H(\state,\Bar{h}))$, where $\mathcal{N}(H)$ represent the nullspace of tangent hyperplane $H$. $H(\state,\Bar{h})$ is the hyperplane intersecting $\state$ and tangent to the obstacle surface described by $\bar{h}$. Denote $\state_i$ as the $i^{\text{th}}$ element in robot state $\state$, $i \in \{1, 2, .., d\}$,
\begin{equation}
    \bm{e}^{(0)} \!=\! [e^{(0)}_1, \ldots, e^{(0)}_d]^\top,\, \text{and} \,
 e^{(0)}_i = 0,\, \text{if} \,\, \partial \Bar{h} / \partial \state_i = 0.  \label{eq:e requirement}
\end{equation}

A geodesic approximation method can then utilize a first-order approximation of the obstacle surface to generate paths $X = \{\state^{(0)}, \state^{(1)}, \dots, \state^{(N)}\}$, which start along each of the $m$ sampled directions and exit the obstacle on its isosurface, where $N \in \mathbb{N}$ is the iteration number~\cite{onManifoldMod}. The function $\phi(\state, \Bar{h}) \in \mathbb{R}^\nstate$ then selects, among the $m$ uniformly sampled candidate directions, the one with the smallest associated potential $P^{(N)}$ computed using Eq.~\eqref{eq:geo approxi}, where $\beta$ is the step size and $p(\state)$ is a user-defined reward function. In most applications, $p(\state)$ is taken as a weighted combination of the distance from $\state^{(i)}$ to the target and the value of $\Bar{h}(\state^{(i)})$.
\begin{equation}
\label{eq:geo approxi}
\begin{aligned}
\state^{(i+1)} &= \beta H(\state^{(i)},\Bar{h})H(\state^{(i)},\Bar{h})^\top \bm{e}^{(i)}+\state^{(i)}, \\
\bm{e}^{(i+1)} &= \frac{H(\state^{(i)},\Bar{h})H(\state^{(i)},\Bar{h})^\top \bm{e}^{(i)}}{||H(\state^{(i)},\Bar{h})H(\state^{(i)},\Bar{h})^\top \bm{e}^{(i)}||_2}, \\
P^{(i+1)} &= P^{(i)} + \beta p(\state^{(i+1)}).
\end{aligned}
\end{equation}

\paragraph{Autonomous Parameter Selection for Obstacle Exiting}
\label{sec: parameter selection}
Constructing virtual obstacles using EFRSs introduces obstacle deformation. For instance, when a human accelerates or decelerates, their EFRS may expand or shrink drastically. MCBF-QP in~\cite{xue_2025_mcbf} selects the exit tangent vector $\phi(\state,\Bar{h})$ using manually tuned constant parameters for iteration number $N$ and step size $\beta$. However, using constant $N$ and $\beta$ causes over-wrapping and under-wrapping issues and outputs inefficient exit strategies $\phi(\state, \Bar{h})$ during the shrinkage or expansion of the EFRS, as shown in Fig.~\ref{fig:wrap sim rep}. 
What's worse, if the exit strategies selected switch frequently between opposite directions due to improper wrapping, we risk introducing local minima that MCBF-QP intends to eliminate back into the trajectories planned. Therefore, to resolve the improper wrapping issue, we propose the autonomous parameter-selection scheme (Algorithm 1) that adapts in real time as the EFRS deforms. The algorithm is motivated by the observation that proper wrapping is achieved whenever $\beta N \!\approx\! d_\text{geo}$, where $d_\text{geo}$ is the geodesic distance between the robot position $\bm{\xi}_\text{rob}$ and the target position $\bm{\xi}_\text{tar}$. This distance is defined as the length of the shortest path lying entirely on the isosurface $\{\bm{\xi} \,|\, \bar{s}(\bm{\xi}) \!=\! \bar{s}(\bm{\xi}_\text{rob})\}$ and connects their surface projections $\bm{\xi}^{\text{pr}}_\text{rob}$ and  $\bm{\xi}^{\text{pr}}_\text{tar}$. Given the correlation between $\beta$ and $N$, fixing one while updating the other via approximated $d_\text{geo}$ trivially ensures proper wrapping. 
In our framework, $N$ is fixed since a static iteration count enables JIT compilation of the geodesic-approximation step, yielding substantial computational speedups.

\begin{algorithm}[t]
    \SetKwInOut{Input}{Input}
    \SetKwInOut{Output}{Output}
    \underline{function AutoParamSel} ($\bar{s}$, $N$, $\bm{\xi}_\text{rob}$, $\bm{\xi}_\text{tar}$)\;
    \textbf{Initialize} empty distance map $\mathcal{M}$. \\
    {\small
    $\mathcal{B}_{\text{all}} = \text{CONTOUR}(\text{EDGE}(\bar{s}(\mathcal{M}), \bar{s}(\bm{\xi}_\text{rob}))$\;
    $d_{\min}=\infty$\;
    \For{$\mathcal{B}\in\mathcal{B}_{\text{all}}$}{
        $\bm{\xi}^{\text{pr}}_\text{rob}, \bm{\xi}^{\text{pr}}_\text{tar} = \text{proj}((\bm{\xi}_\text{rob}, \bm{\xi}_\text{tar}),\mathcal{B})$\;
        \If{$\|\bm{\xi}_\text{rob}-\bm{\xi}^{\text{pr}}_\text{rob}\|_2<d_{\min}$}{
            $d_{\min}=||\bm{\xi}_\text{rob}-\bm{\xi}^{\text{pr}}_\text{rob}||$\;
            $d_{\text{geo}} = \text{GEO}(\mathcal{B}, \bm{\xi}^{\text{pr}}_\text{rob}, \bm{\xi}^{\text{pr}}_\text{tar})$\;
        }
    }
    \textbf{return} $\beta=d_{\text{geo}} / N$
    }
    \caption{Autonomous parameter selection}\label{alg:autoparam}
\end{algorithm}

Input arguments to the algorithm include the combined DF $\bar{s}$, iteration number $N$, robot position $\bm{\xi_{\text{rob}}}$, and target position $\bm{\xi_{\text{tar}}}$. Firstly, the algorithm initializes an empty distance map based on the detectable obstacle set $Q$ and a pre-defined resolution. The distance map is then filled with the distances to obstacles computed using combined DF $\bar{s}$:
\begin{equation}
\label{eq:merged df}
    \Bar{s}(\bm{\xi}) = -\frac{1}{\rho}\log\bigg(\sum_{o \in Q} \exp{(-\rho [s_o(\bm{\xi}, \ostate)+1])}\bigg)-1,
\end{equation}
which can be computed given 2D DF $s_o(\bm{\xi}, \ostate)$ for each obstacle $o \in Q$. 
Similar to what we did during EBM postprocessing in Section \ref{sec:barrier function construction}, edge and contour detection is used to extract contours $\mathcal{B}_{\text{all}}$ on the level set $\{\bm{\xi} \,|\, \bar{s}(\bm{\xi}) = \bar{s}(\bm{\xi}_\text{rob})\}$ from the distance map. The above computation can produce multiple disconnected contours, but only one corresponds to the isoline that the geodesic approximator should follow. In Algorithm 1, we iterate through all the extracted contours and set $d_\text{geo}$ to the geodesic distance on the contour that is closest to the robot position $\bm{\xi}_\text{rob}$. The function $\text{proj}(\cdot)$ calculates the projections of given points on the contour. The function $\text{GEO}(\cdot)$ calculates the geodesic distance between two points on a boundary.

\subsection{Proactive and Adaptive MCBF Formulation}
Summarizing the modifications, the optimization problem for the proposed MCBF-QP variant can be formulated as in Eq.~\eqref{eq:adaptive mcbf}. Note that here $\phi(\state,h_o)$ from the standard MCBF-QP constraint \eqref{eq:mod_phi_cbf_constraint_affine} is replaced with $\phi(\state,\bar{h})$ using the combined barrier function. This change enables MCBF-QPs to ensure local-minimum-free circumvention not only for individual concave obstacles but also for concave obstacle unions formed by dynamic obstacles, like humans, with one another, or with static infrastructure in the environment. Eq.~\eqref{eq:cbf-qp h_o safety constraints} is introduced to prevent the robot from entering reentrant corners $Q_g$ in between obstacles, defined as $Q_g = \{\state: h_o(\state,\ostate)>0, \forall o \in O \text{ and } \bar{h}(\state)\leq0\}$. Robot entering $Q_g$ will destabilize the combined DF in \eqref{eq:merged df} and make the parameter selection in Algorithm~\ref{alg:autoparam} inaccurate.
\begin{align}
    \action_{\text{mcbf}} &= \arg\min_{\action \in \mathbb{R}^\naction}\;(\action-\action_\text{nom})^\top(\action-\action_\text{nom}), & \nonumber\\
    \text{s.t.}\quad
    &\lie{f} h_o + \lie{g} h_o \action + \nabla_{\ostate}h_o^\top\dostate \geq - \alpha \left(h_o\right) \, \forall o \in O, \\
     &\lie{f} \bar{h} + \lie{g} \bar{h} \action  \geq - \alpha \left(\bar{h}\right), \label{eq:cbf-qp h_o safety constraints}\\
    &\phi(\state,\bar{h})^\top f(\state)+\phi(\state,\bar{h})^\top g(\state)\action \geq \gamma, \label{eq:adaptive mcbf}
\end{align}
 
\begin{figure}[tb]
    \centering
    \includegraphics[width=\linewidth]{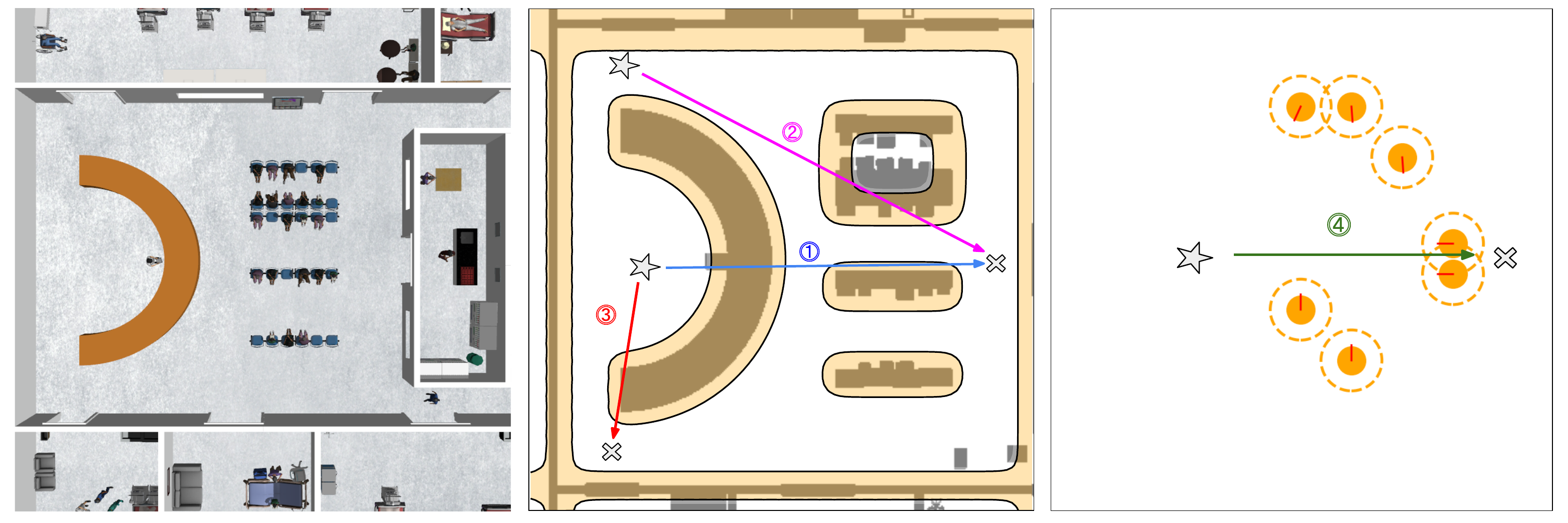}
    \caption{Hospital (left, middle) and crowd (right) simulation environment setup for robot comparison tests, including the Gazebo world (left) and the static environmental layout in the robot's perspective (middle, right).\label{fig:hos sim rep}}
    \vspace{-10pt}
\end{figure}

\section{Experiment}
\label{sec:evaluation}
The proposed framework is evaluated on a differential-drive Fetch robot across kinematic simulations, Gazebo simulations, and real-world experiments\footnote{%
  \parbox[t]{\linewidth}{%
    Code: \url{https://github.com/yifanxueseas/mmp_mcbf_control}\\
    Video: \url{https://youtu.be/heagLEwfcVc}
  }%
}. Four scenarios are designed for kinematic simulations (middle and right images of Fig.~\ref{fig:hos sim rep}), featuring hospital and crowd navigation tasks. The robot is required to move from the star-marked initial position to the cross-marked target, under 10 uniformly sampled initial orientations. Each algorithm’s performance is assessed in terms of safety, efficiency, and solver feasibility, as reported in Table~\ref{tab:hos_eva}. The performance of the proposed methods and baselines in hospital environments is further validated in Gazebo simulations (left image of Fig.~\ref{fig:hos sim rep}). In both kinematic and Gazebo hospital simulations, i.e., Scenarios 1 to 3, the EBM-based motion predictor is used; while in the crowd navigation task (Scenario 4) and the real-world experiments, the CVM-based motion predictor is used.

\textbf{Hardware, Computation, and Parameters}: All runtime is recorded on an Intel Core™ i7-12700K × 20 with NVIDIA GeForce RTX 3090 Ti (only for EBMs). The adaptive MCBF loops at 30 Hz while the whole framework runs at 10 Hz, due to all components running in sequence on the same computer. In practice, they can run in different compute nodes, so that the controller can be sped up to the runtime of standard CBF-QP (about 100 Hz).

As in \cite{ze_2025_ebm}, the EBM-based motion predictor has a classic UNet architecture \cite{ronneberger_2015_unet}, and is trained on a synthetic dataset collected from the hospital scenario, containing about 30K data samples from 630 trajectories. Each trajectory is from a simulated pedestrian with different motion speeds and random motion noise. This motion noise can be regarded as measurement noise or staggering behaviors in reality. The horizon for motion prediction is 20 steps (4 s). The GPDF uses the Matérn $\nu=1/2$ kernel: $k_o(d)=\sigma^2\exp{-d/L}$ with $\sigma=1$ and $L=0.2$. Because GPDF training scales as $O(n^3)$ \cite{gpdf}, we capped each obstacle’s boundary representation at 200 points during online barrier-function generation. This coarser sampling inevitably smooths small gaps and fine details, but we found that such smoothing is beneficial: it prevents the controller from attempting to navigate through narrow, impractical openings and thus improves overall navigation safety, while still preserving the essential obstacle geometry needed for the robot to plan feasible paths.

\textbf{Robot Dynamics}: The Fetch robot is modeled as a unicycle whose point of interest is $a$ meters ahead of the wheel axis, representing an output transformation of the default model. The model is referred to as \emph{shifted} when $a\!>\!0$, and as \emph{standard} when $a\!=\!0$. The resulting control-affine dynamics are
\begin{equation}
\label{eq:shifted unicycle}
    \dot{\state}=\begin{bmatrix}
        \dot{p}_x\\\dot{p}_y\\\dot{\theta}
    \end{bmatrix}=
    \begin{bmatrix}
        \cos{\theta} & -a\sin{\theta}\\ \sin{\theta} & a\cos{\theta}\\ 0 & 1
    \end{bmatrix}
    \begin{bmatrix}
        v \\ \omega
    \end{bmatrix}.
\end{equation}
For CBF-QPs and MCBF-QPs used in our experiments and the 2D examples in Fig.~\ref{fig:wrap sim rep}, we employ the shifted model with $a\!=\!0.2 \;\text{m}$. This transformation is advantageous because the Euclidean DF $s_o(\bm{\xi}, \ostate)$ to the robot depends directly on the control inputs $(v,\omega)$ through their first time derivative, enabling its direct use as CBF without auxiliary dynamics or differentiation.
However, not all underactuated control-affine systems admit such a shift; many require DF augmentation to construct relative-degree-one CBFs. Therefore, we further evaluate the proposed autonomous parameter selection algorithm on the standard unicycle model ($a=0$) with augmented CBFs (A-CBFs) in the 3D examples of Fig.~\ref{fig:wrap sim rep}, demonstrating the framework’s theoretical applicability to general control-affine systems.

\textbf{Barrier Function Designs}: In unicycle kinematics, $\bm{\xi}=[p_x, p_y]^\top$ and $\state=[p_x, p_y, \theta]^\top$. Thus, the barrier function $h_o$ for the shifted unicycle model can be adopted from $s_o$ as
\begin{equation}
\label{eq: 2d barrier}
    h_o(\state,\ostate) = h_o(\bm{\xi}, \ostate) = s_o(\bm{\xi}, \ostate).
\end{equation}
Since the partial derivatives of $s_o$ with respect to $\theta$ are all zeros, the corresponding CBF constraints can be written as
\begin{align*}
    \nabla_{\bm{\xi}} h_o(\bm{\xi},\ostate)^\top \dot{\bm{\xi}} +\nabla_{\ostate}h_o(\bm{\xi},\ostate)^\top\dot{\bm{x}}_o \geq -\alpha (h_o(\bm{\xi},\ostate)).
\end{align*}

For the standard model, we formulate our adaptive MCBF-QP using A-CBF $h_\text{aug}(\state,\ostate)$ as follows, where $w \in \mathbb{R}^+$ \cite{xue_2025_mcbf}.
\begin{equation}
\begin{aligned}
h_\text{aug}(\state, \ostate) = s_o(\bm{\xi},\ostate) + w\nabla_{\bm{\xi}}s_o(\bm{\xi},\ostate)^\top\begin{bmatrix}
 \cos{\theta} \\
 \sin{\theta}
\end{bmatrix}
    \label{eq:ho barrier function}   
\end{aligned}
\end{equation}
Note that the geodesic approximation of A-CBFs is performed in a high-dimensional non-Euclidean space, as visualized by the 3D cases in Fig. \ref{fig:wrap sim rep}. A-CBF constraints in CBF-QP and MCBF-QP can be obtained by substituting $h_o = h_\text{aug}(x, \ostate)$  directly into the CBF constraint inequality in Eq.~\eqref{eq:cbf-qp safety constraints}, and the combined barrier function computation in \eqref{eq:uni h}.

\textbf{Nominal controller}: In all experiments and figures in this paper, the nominal controllers are approximated from the linear dynamical systems $\bm{u}_\text{nom}^l=\frac{\bm{\xi}^*-\bm{\xi}}{||\bm{\xi}^*-\bm{\xi}||}$ as in \eqref{eq:dubin nominal system}, where $\bm{\xi}$ is the robot position, $\bm{\xi}^*$ is the target position, $\psi$ is the angle difference between the current robot pose $\theta$ and the desired robot pose estimated using the orientation of $\bm{u}_\text{nom}^l$, and $\Delta t$ is the sampling time of the nominal controller. The approximation is not exact due to the non-holonomic property of differential drive systems. 
\begin{equation}
    v_\text{nom} = ||u_\text{nom}^l||_2, \quad \omega_\text{nom} = \psi / \Delta t \label{eq:dubin nominal system}
\end{equation}

\begin{table*}[t]
\caption{Evaluation of different methods in hospital (Scenarios 1-3) and social navigation (Scenario 4). Ten runs were conducted for each scenario.}
\label{tab:hos_eva}
\centering
\small
\setlength{\tabcolsep}{8pt}
\renewcommand{\arraystretch}{1.2}

\begin{tabular}{lccccc c}
\toprule
 & \multicolumn{2}{c}{\textbf{Runtime (s)}} &
 \multicolumn{3}{c}{\textbf{Scenarios 1 / 2 / 3 / 4}} &
 \textbf{Timespan (s)} \\
\cmidrule(lr){2-3} \cmidrule(lr){4-6}
\textbf{Method} & Mean & Std &
 Safety \# & Success \# & Infeasibility \# & Mean \\
\midrule

MPPI & 0.028 & 0.011 &  10 / 10 / 0 / 10 &  0 / 0 / 9 / 10  &  0 / 0 / 0 / 0 & -- / -- / 18.4 / 23.7 \\ 
CBF       & 0.010 & 0.009 & 10 / 10 / 10 / 2 & 0 / 10 / 9 / 10 &   0 / 1 / 0 / 69    & -- / 28.9 / 25.0 / 16.3 \\
MCBF      & 0.018 & 0.009 & 2 / 5 / 4 / 0   & 10 / 4 / 10 / 10 &  59 / 426 / 42 / 122 & 25.0 / 19.3 / 13.2 / 13.4 \\
MMP-MPC   & 0.066 & 0.046 &  3 / 0 / 0 / 0  & 0 / 1 / 6 / 4  &   0 / 0 / 0 / 0     & -- / 33.6 / 12.5 / 14.2 \\
MMP-MPPI & 0.045 & 0.005 & 10 / 10 / 10 / 10 & 0 / 0 / 10 / 10 &  0 / 0 / 0 / 0 & -- / -- / 16.1 / 22.5 \\
MMP-CBF   & 0.010 & 0.007 & 10 / 10 / 1 / 10 & 1 / 5 / 9 / 10  &   0 / 317 / 77 / 0  & 26.1 / 34.0 / 12.9 / 9.8 \\
\midrule
MMP-MCBF (ours) & 0.036 & 0.018 & 10 / 10 / 10 / 10 & 10 / 10 / 10 / 10 & 10 / 0 / 0 / 0 & 31.0 / 19.3 / 15.5 / 9.6 \\
\bottomrule
\end{tabular}
\end{table*}

\textbf{Metrics}: The evaluation in \autoref{tab:hos_eva} contains the following aspects: ``Safety \#'' and ``Success \#'' show the number of cases where the robot did not collide with obstacles, and reached the assigned goals out of ten runs in each scenario. ``Infeasibility \#'' denotes the number of time steps per trajectory (out of 800) in which the optimization solver fails to find a safety-feasible solution. ``Timespan'' is the average duration of the goal-reached trajectories.

\begin{figure*}[t]
    \centering
    \begin{subfigure}{0.245\textwidth}
         \includegraphics[width=\textwidth]{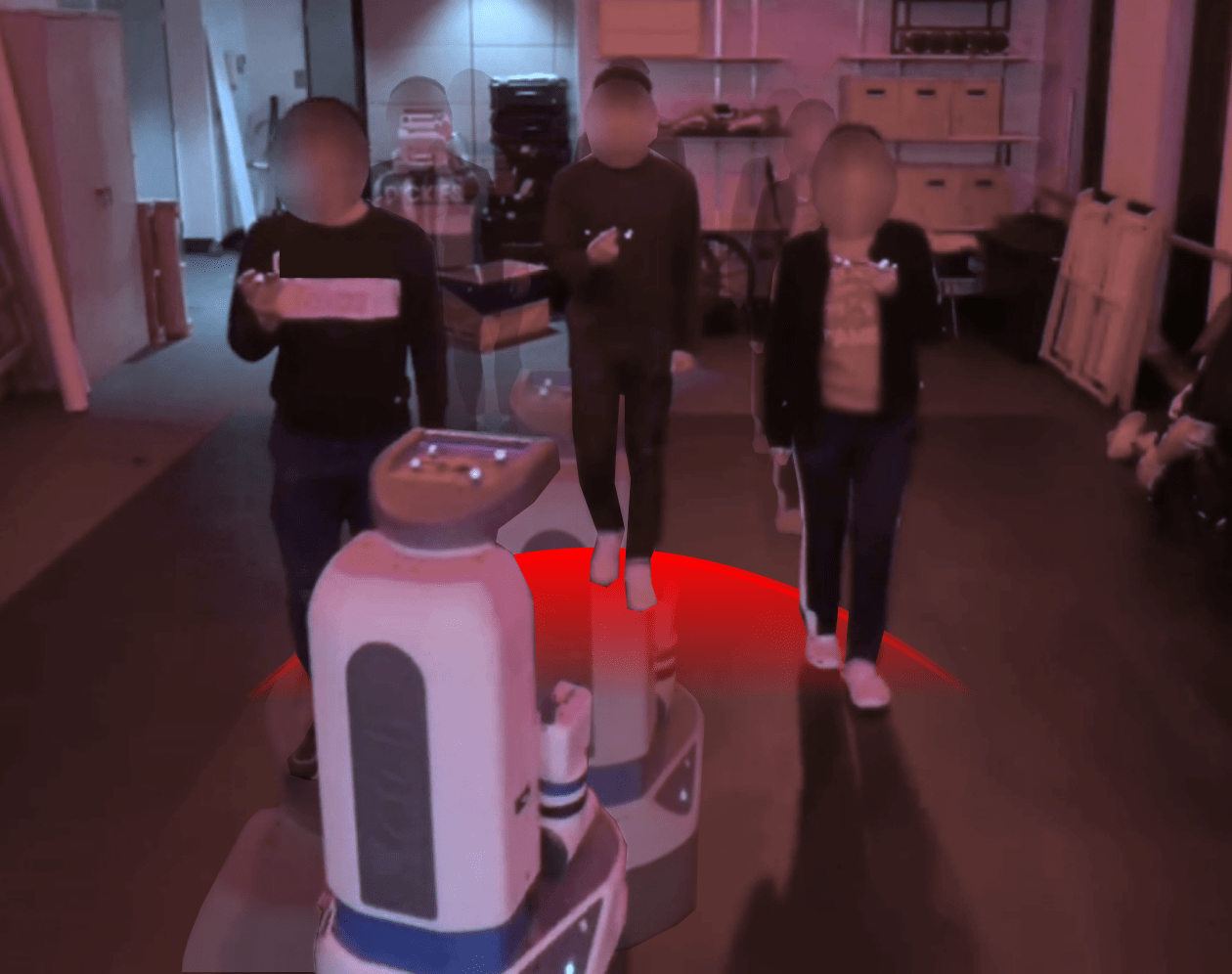}
    \caption{MPC, with prediction\label{fig:fv_mpc_pred}}
    \end{subfigure}
    \begin{subfigure}{0.245\textwidth}
         \includegraphics[width=\textwidth]{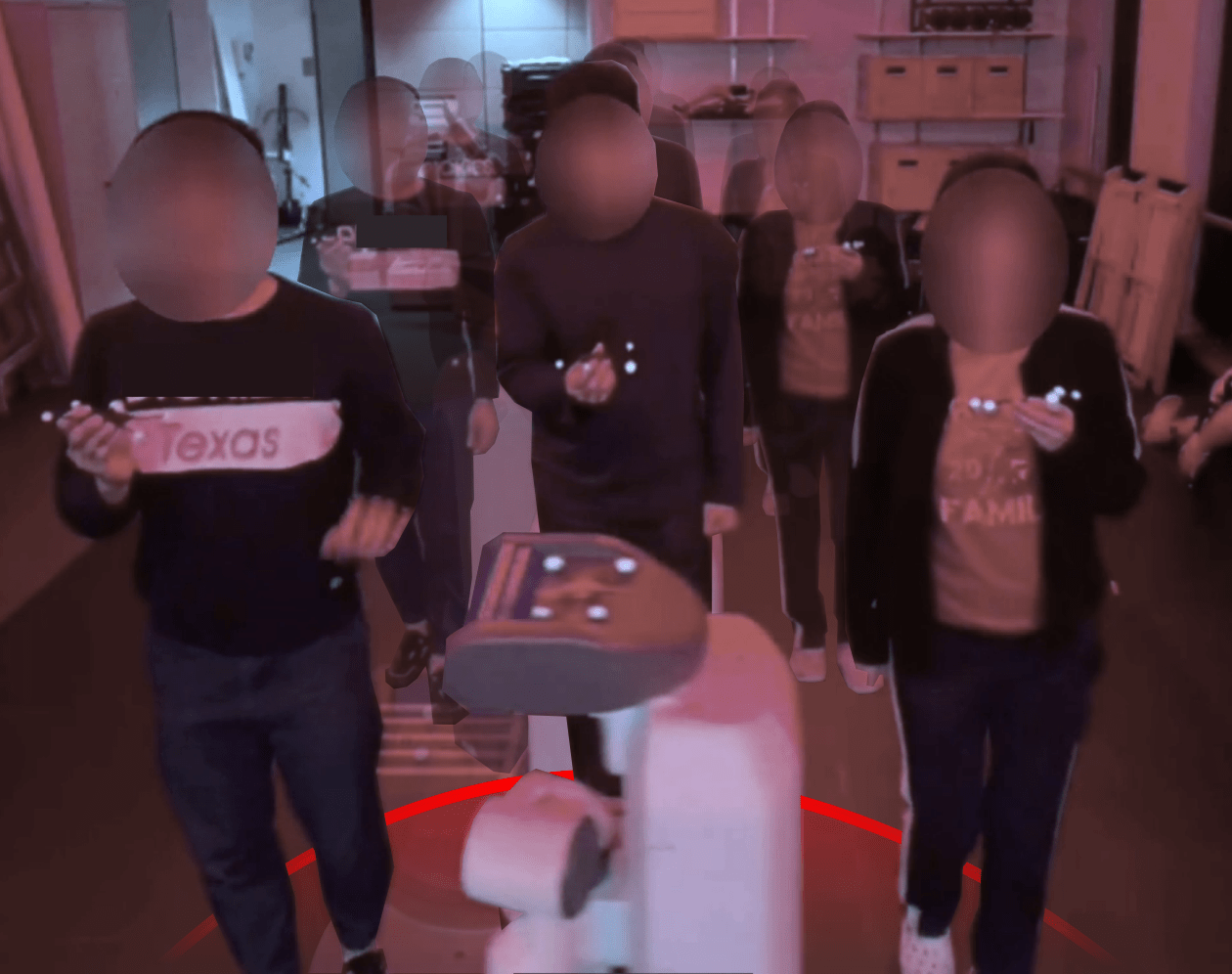}
     \caption{CBF, with prediction\label{fig:fv_cbf_pred}}
    \end{subfigure}
    \begin{subfigure}{0.245\textwidth}
         \includegraphics[width=\textwidth]{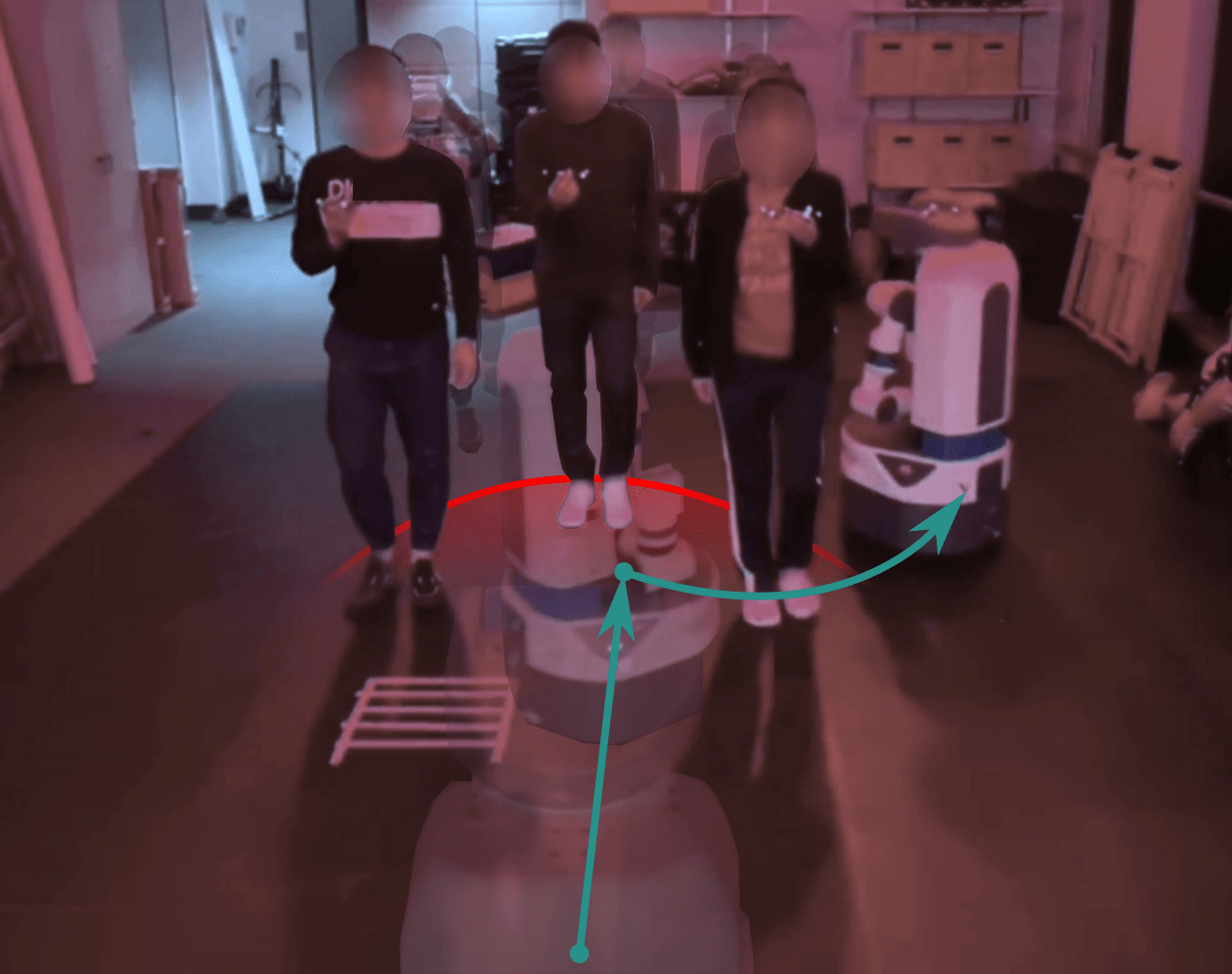}
     \caption{MCBF, no prediction\label{fig:fv_mcbf_nopred}}
    \end{subfigure}
    \begin{subfigure}{0.245\textwidth}
         \includegraphics[width=\textwidth]{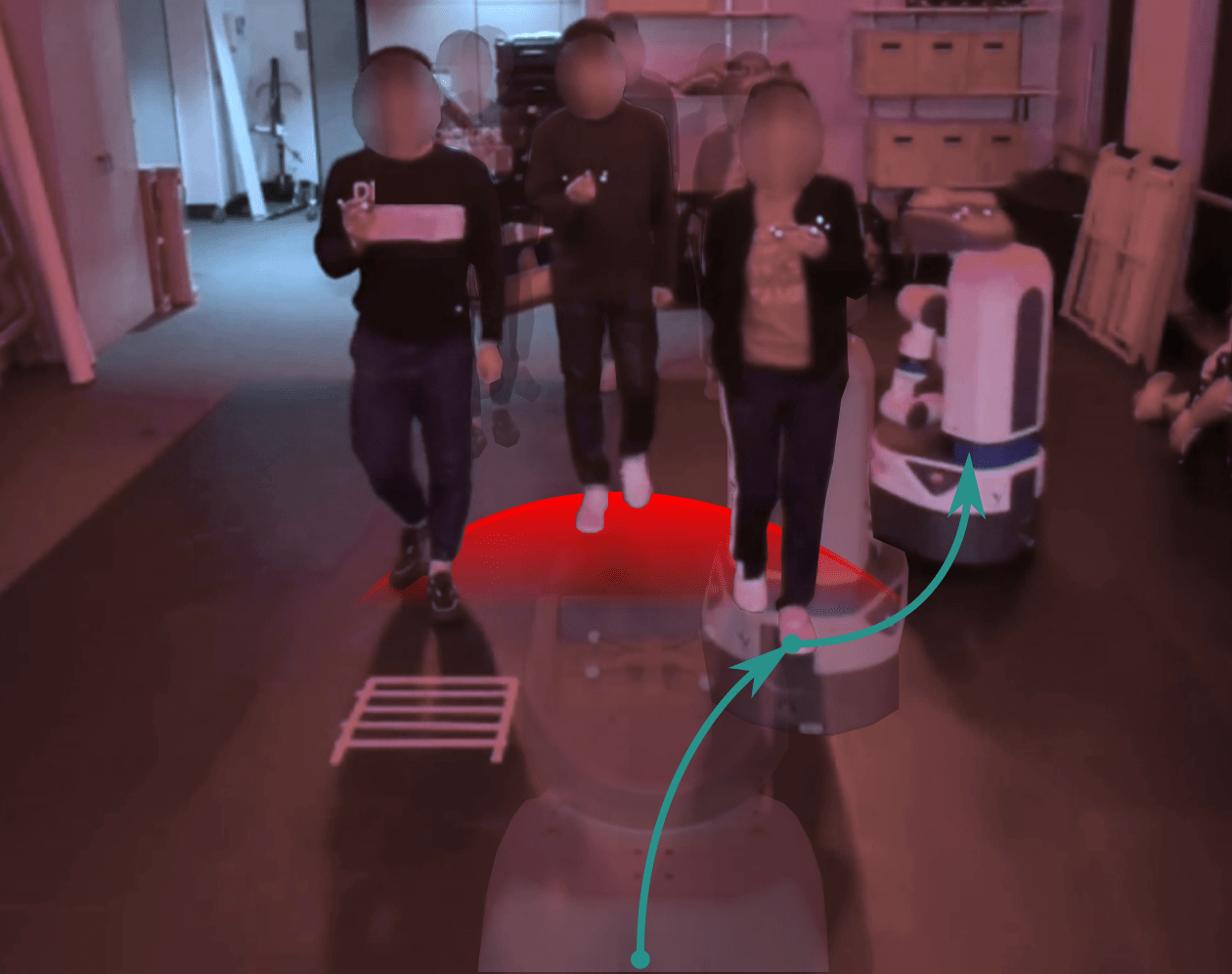}
    \caption{MCBF, with prediction\label{fig:fv_mcbf_pred}}
    \end{subfigure}
    \caption{Real-world experiments. A Fetch robot avoids three pedestrians approaching in a U-shape formation.
    }
    \label{fig:real_exp}
\end{figure*}

\textbf{Result Analysis}: The proposed approach is evaluated against existing safe local planners, including MPC, MPPI, CBF-QP, and MCBF-QP. As an ablation study, we compare them with and without motion prediction.
For real-time performance, the penalty-based MPC is constrained to 0.1s in all experiments. As shown in \autoref{tab:hos_eva}, pairing motion prediction with the adaptive MCBF-QP yields collision-free navigation with high target-reaching rates in dense dynamic concave environments. This integration substantially improves the baseline MCBF-QP by preventing unsafe attempts to pass through narrow gaps when humans approach. Although integrating motion prediction with MPC, MPPI, and CBF-QP increases their safety margins, they usually fail to generate alternative safe routes without the exit-force guidance inherent to MCBF-QP. While MPC can succeed in concave environments~\cite{obca}, doing so typically requires careful initialization via computationally expensive global planners. In terms of efficiency, MMP-MCBF yields the shortest average collision-free travel time. MMP-CBF and MCBF-QP produce slightly shorter paths in Scenarios 1 and 3, but at the cost of higher collision rates or reduced target-reaching success.
We further validated the effectiveness of our proposed framework using real-world crowd navigation experiments, as shown in Fig.~\ref{fig:real_exp}. The hardware experiments reinforce our observations from simulations that MMP-MCBF can circumvent obstacles more proactively than standard MCBF-QPs and CBF-QPs, and induce fewer robot freezing points than MMP-MPC and MMP-CBF. 

\textbf{Discussion}: In the current algorithm, we formulate the robot's safety as a continuous-time deterministic optimization problem. In comparison to standard CBF-QPs, introducing EFRS as obstacles shrinks the safe regions for the robot and can subject the controller to solver infeasibility issues if the prediction length value $\tau_{\max}$ is too large. 
In the experiments, the prediction horizon $\tau_{\max}$ is set to 4 seconds to balance reactivity and safety: a longer horizon may lead to overly conservative behavior and freezing robots, while a shorter horizon may provide insufficient time for the robot to react.
During the experiments, short-term abnormal noise and occasional unusual motion behaviors do not lead to collisions or solver failure, and the robot can continue smoothly afterwards. However, long-term anomalies—such as sensor failure or intentional obstruction—can invalidate motion prediction and therefore require dedicated anomaly detection mechanisms and additional safety modules.
Future work on reformulating the problem using a stochastic system or enabling online adaptation of prediction length $\tau_{\max}$ could alleviate infeasibility issues and offer better navigation efficiency.
Additionally, the robot is given access to the full obstacle geometry once an obstacle enters the detection range, which might be unrealistic due to limited sensor field of view, occlusions, and incomplete perception. Since the autonomous parameter selection algorithm provides a foundation for MCBF-QP in environments with dynamically deforming obstacles, the proposed controller has the potential to adapt online to partial and evolving obstacle information by treating gradually observed obstacles as deforming. A natural future work is to integrate this framework with a real perception pipeline that incrementally constructs obstacle representations from partial observations.

\section{Conclusion}
\label{sec:conclusion}
In this work, we proposed a safe and efficient mobile robot navigation pipeline, named MMP-MCBF, based on the integration of motion predictions of dynamic obstacles and an improved on-manifold control barrier function method, which enables mobile robots to avoid dynamic obstacles proactively. To consider motion prediction in the control barrier function framework, we further improved the on-manifold MCBF-QP algorithm for adaptive and autonomous parameter searching. Apart from the traditional CVM-based prediction, a learning-based EBM motion predictor is investigated in this pipeline to consider multimodal motion prediction for more comprehensive and flexible prediction. From both simulated scenarios and real-world experiments, the proposed MMP-MCBF approach outperforms other popular obstacle avoidance approaches, shown to be safe and efficient in complex and dynamic environments, especially with the presence of non-convex static or dynamic obstacles.

\bibliographystyle{IEEEtran}
\bibliography{main}

@article{mavrogiannis_2023_socialnavi,
  title={Core challenges of social robot navigation: A survey},
  author={Mavrogiannis, Christoforos and Baldini, Francesca and Wang, Allan and Zhao, Dapeng and Trautman, Pete and Steinfeld, Aaron and Oh, Jean},
  journal={ACM Transactions on Human-Robot Interaction},
  volume={12},
  number={3},
  pages={1--39},
  year={2023},
  publisher={ACM New York, NY}
}

@book{correll_2022_introduction,
  title={Introduction to autonomous robots: mechanisms, sensors, actuators, and algorithms},
  author={Correll, Nikolaus and Hayes, Bradley and Heckman, Christoffer and Roncone, Alessandro},
  year={2022},
  publisher={MIT Press}
}

@book{billard_2022_learning,
  title={Learning for adaptive and reactive robot control: a dynamical systems approach},
  author={Billard, Aude and Mirrazavi, Sina and Figueroa, Nadia},
  year={2022},
  publisher={MIT Press}
}

@inproceedings{ames_2019_cbf,
  title={Control barrier functions: Theory and applications},
  author={Ames, Aaron D and Coogan, Samuel and Egerstedt, Magnus and Notomista, Gennaro and Sreenath, Koushil and Tabuada, Paulo},
  booktitle={European control conference (ECC)},
  pages={3420--3431},
  year={2019},
  organization={IEEE}
}

@INPROCEEDINGS{cosner2023robust,
  author={Cosner, Ryan K and Culbertson, Preston and Taylor, Andrew J and Ames, Aaron D},
  booktitle={Robotics: Science and Systems (RSS)}, 
  title={Robust safety under stochastic uncertainty with discrete-time control barrier functions}, 
  pages={1-11},
  year={2023},
}

@ARTICLE{adaptivecbf,
  author={Xiao, Wei and Belta, Calin and Cassandras, Christos G.},
  journal={IEEE Transactions on Automatic Control}, 
  title={Adaptive Control Barrier Functions}, 
  year={2022},
  volume={67},
  number={5},
  pages={2267-2281},
  doi={10.1109/TAC.2021.3074895}
}

@article{huber_2022_avoiding,
  title={Avoiding dense and dynamic obstacles in enclosed spaces: Application to moving in crowds},
  author={Huber, Lukas and Slotine, Jean-Jacques and Billard, Aude},
  journal={IEEE Transactions on Robotics},
  volume={38},
  number={5},
  pages={3113--3132},
  year={2022},
  publisher={IEEE}
}

@ARTICLE{onManifoldMod,
  author={Fourie, Christopher K. and Figueroa, Nadia and Shah, Julie A.},
  journal={IEEE Transactions on Robotics}, 
  title={On-Manifold Strategies for Reactive Dynamical System Modulation With Nonconvex Obstacles}, 
  year={2024},
  volume={40},
  number={},
  pages={2390-2409},
  doi={10.1109/TRO.2024.3378179}
}

@article{xue_2025_mcbf,
  title={No Minima, No Collisions: Combining Modulation and Control Barrier Function Strategies for Feasible Dynamical Collision Avoidance}, 
  author={Xue, Yifan and Figueroa, Nadia},
  journal={arXiv preprint:2502.14238},
  year={2025}
}

@INPROCEEDINGS{prednomcbf,
  author={Breeden, Joseph and Panagou, Dimitra},
  booktitle={Conference on Decision and Control (CDC)}, 
  title={Predictive Control Barrier Functions for Online Safety Critical Control}, 
  year={2022},
  pages={924-931},
  publisher={IEEE},
  doi={10.1109/CDC51059.2022.9992926}
}

@INPROCEEDINGS{predstatecbf,
  author={Jian, Zhuozhu and Yan, Zihong and Lei, Xuanang and Lu, Zihong and Lan, Bin and Wang, Xueqian and Liang, Bin},
  booktitle={ICRA}, 
  title={Dynamic Control Barrier Function-based Model Predictive Control to Safety-Critical Obstacle-Avoidance of Mobile Robot}, 
  year={2023},
  pages={3679-3685},
  publisher={IEEE},
  doi={10.1109/ICRA48891.2023.10160857}
}

@ARTICLE{zhang_2023_wtampc,
  author={Zhang, Ze and Hajieghrary, Hadi and Dean, Emmanuel and Åkesson, Knut},
  journal={IEEE Robotics and Automation Letters}, 
  title={Prescient Collision-Free Navigation of Mobile Robots With Iterative Multimodal Motion Prediction of Dynamic Obstacles}, 
  year={2023},
  volume={8},
  number={9},
  pages={5488-5495},
  doi={10.1109/LRA.2023.3296333}
}

@INPROCEEDINGS{Heuer_2023_proactive,
  title={Proactive Model Predictive Control with Multi-Modal Human Motion Prediction in Cluttered Dynamic Environments}, 
  author={Heuer, Lukas and Palmieri, Luigi and Rudenko, Andrey and Mannucci, Anna and Magnusson, Martin and Arras, Kai O.},
  booktitle={IROS}, 
  year={2023},
  publisher={IEEE},
  pages={229-236},
}

@article{samavi_2024_sicnav,
  title={{SICN}av: Safe and Interactive Crowd Navigation using Model Predictive Control and Bilevel Optimization},
  author={Samavi, Sepehr and Han, James R and Shkurti, Florian and Schoellig, Angela P},
  journal={IEEE Transactions on Robotics},
  year={2024},
  publisher={IEEE}
}

@INPROCEEDINGS{mppi,
  author={Williams, Grady and Drews, Paul and Goldfain, Brian and Rehg, James M. and Theodorou, Evangelos A.},
  booktitle={2016 IEEE International Conference on Robotics and Automation (ICRA)}, 
  title={Aggressive driving with model predictive path integral control}, 
  year={2016},
  volume={},
  number={},
  pages={1433-1440},
  doi={10.1109/ICRA.2016.7487277}}

@INPROCEEDINGS{mpccbf,
  author={Zeng, Jun and Zhang, Bike and Sreenath, Koushil},
  booktitle={American Control Conference (ACC)}, 
  title={Safety-Critical Model Predictive Control with Discrete-Time Control Barrier Function}, 
  year={2021},
  volume={},
  number={},
  pages={3882-3889},
  doi={10.23919/ACC50511.2021.9483029}
}

@INPROCEEDINGS{mpccbflayered,
  author={Grandia, Ruben and Taylor, Andrew J. and Ames, Aaron D. and Hutter, Marco},
  booktitle={ICRA}, 
  title={Multi-Layered Safety for Legged Robots via Control Barrier Functions and Model Predictive Control}, 
  year={2021},
  volume={},
  number={},
  pages={8352-8358},
  publisher={IEEE},
  doi={10.1109/ICRA48506.2021.9561510}
}

@INPROCEEDINGS{tai_2017_sim2real,
  author={Tai, Lei and Paolo, Giuseppe and Liu, Ming},
  booktitle={IROS}, 
  title={Virtual-to-real deep reinforcement learning: Continuous control of mobile robots for mapless navigation}, 
  year={2017},
  pages={31-36},
  publisher={IEEE},
  doi={10.1109/IROS.2017.8202134}
}

@INPROCEEDINGS{ceder_2024_ddpgmpc,
  author={Ceder, Kristian and Zhang, Ze and Burman, Adam and Kuangaliyev, Ilya and Mattsson, Krister and Nyman, Gabriel and Petersén, Arvid and Wisell, Lukas and Åkesson, Knut},
  booktitle={IROS}, 
  title={Bird’s-Eye-View Trajectory Planning of Multiple Robots using Continuous Deep Reinforcement Learning and Model Predictive Control}, 
  year={2024},
  pages={8002-8008},
  publisher={IEEE},
  doi={10.1109/IROS58592.2024.10801434}
}

@INPROCEEDINGS{Xu_2023_drllimit,
  author={Xu, Zifan and Liu, Bo and Xiao, Xuesu and Nair, Anirudh and Stone, Peter},
  booktitle={ICRA}, 
  title={Benchmarking Reinforcement Learning Techniques for Autonomous Navigation}, 
  year={2023},
  pages={9224-9230},
  publisher={IEEE},
  doi={10.1109/ICRA48891.2023.10160583}
}

@article{scholler_2020_cvm,
  title={What the constant velocity model can teach us about pedestrian motion prediction},
  author={Sch{\"o}ller, Christoph and Aravantinos, Vincent and Lay, Florian and Knoll, Alois},
  journal={IEEE Robotics and Automation Letters},
  volume={5},
  number={2},
  pages={1696--1703},
  year={2020},
  publisher={IEEE}
}

@ARTICLE{ze_2025_ebm,
    title={Future-Oriented Navigation: Dynamic Obstacle Avoidance with One-Shot Energy-Based Multimodal Motion Prediction}, 
    author={Ze Zhang and Georg Hess and Junjie Hu and Emmanuel Dean and Lennart Svensson and Knut Åkesson},
    journal={IEEE Robotics and Automation Letters}, 
    year={2025},
    pages={1-8},
    doi={10.1109/LRA.2025.3575969}
}

@inproceedings{oleynikova_2016_signed,
  title={Signed distance fields: A natural representation for both mapping and planning},
  author={Oleynikova, Helen and Millane, Alexander and Taylor, Zachary and Galceran, Enric and Nieto, Juan and Siegwart, Roland},
  booktitle={RSS workshop: geometry and beyond-representations, physics, and scene understanding for robotics},
  year={2016},
}

@article{lecun_2006,
  title={A tutorial on energy-based learning},
  author={LeCun, Yann and Chopra, Sumit and Hadsell, Raia and Ranzato, M and Huang, Fujie and others},
  journal={Predicting structured data},
  year={2006}
}

@INPROCEEDINGS{gpdf,
  author={Choi, Ho Jin and Figueroa, Nadia},
  booktitle={ICRA}, 
  title={Towards Feasible Dynamic Grasping: Leveraging Gaussian Process Distance Field, SE(3) Equivariance, and Riemannian Mixture Models}, 
  year={2024},
  pages={6455-6461},
  publisher={IEEE},
  doi={10.1109/ICRA57147.2024.10611601}
}

@article{le_2023_gpdf,
  title={Accurate {G}aussian-process-based distance fields with applications to echolocation and mapping},
  author={Le Gentil, Cedric and Ouabi, Othmane-Latif and Wu, Lan and Pradalier, Cedric and Vidal-Calleja, Teresa},
  journal={IEEE Robotics and Automation Letters},
  volume={9},
  number={2},
  pages={1365--1372},
  year={2023},
  publisher={IEEE}
}

@ARTICLE{LukesDS,
  author={Huber, Lukas and Billard, Aude and Slotine, Jean-Jacques},
  journal={IEEE Robotics and Automation Letters}, 
  title={Avoidance of Convex and Concave Obstacles With Convergence Ensured Through Contraction}, 
  year={2019},
  volume={4},
  number={2},
  pages={1462-1469},
  doi={10.1109/LRA.2019.2893676}
}

@ARTICLE{clfcbfEquilibria,
  author={Reis, Matheus F. and Aguiar, A. Pedro and Tabuada, Paulo},
  journal={IEEE Control Systems Letters}, 
  title={Control Barrier Function-Based Quadratic Programs Introduce Undesirable Asymptotically Stable Equilibria}, 
  year={2021},
  volume={5},
  number={2},
  pages={731-736},
  doi={10.1109/LCSYS.2020.3004797}
}

@article{rios_2015_proxemics,
  title={From proxemics theory to socially-aware navigation: A survey},
  author={Rios-Martinez, Jorge and Spalanzani, Anne and Laugier, Christian},
  journal={International Journal of Social Robotics},
  volume={7},
  pages={137--153},
  year={2015},
  publisher={Springer}
}

@inproceedings{ronneberger_2015_unet,
  title={{U}-{N}et: Convolutional networks for biomedical image segmentation},
  author={Ronneberger, Olaf and Fischer, Philipp and Brox, Thomas},
  booktitle={Medical image computing and computer-assisted intervention},
  pages={234--241},
  year={2015},
  organization={Springer}
}

@ARTICLE{obca,
  author={Zhang, Xiaojing and Liniger, Alexander and Borrelli, Francesco},
  journal={IEEE Transactions on Control Systems Technology}, 
  title={Optimization-Based Collision Avoidance}, 
  year={2021},
  volume={29},
  number={3},
  pages={972-983},
  doi={10.1109/TCST.2019.2949540}
}

\end{document}